\documentclass{article}

\usepackage[final]{neurips2024_ws/neurips_2024}

\usepackage[utf8]{inputenc} % allow utf-8 input
\usepackage[T1]{fontenc}    % use 8-bit T1 fonts
\usepackage{hyperref}       % hyperlinks
\usepackage{url}            % simple URL typesetting
\usepackage{booktabs}       % professional-quality tables
\usepackage{amsfonts}       % blackboard math symbols
\usepackage{nicefrac}       % compact symbols for 1/2, etc.
\usepackage{microtype}      % microtypography
\usepackage{xcolor}         % colors

\usepackage{todonotes}
\newcommand{\note}[1]{\todo[inline]{#1}}
\renewcommand{\note}[1]{}  % Uncomment to remove all notes, e.g., to judge page count or to submit
\usepackage{amsmath,amsfonts,bm}

\def\eqref#1{equation~\ref{#1}}
\def\1{\bm{1}}

\DeclareMathAlphabet{\mathsfit}{\encodingdefault}{\sfdefault}{m}{sl}
\SetMathAlphabet{\mathsfit}{bold}{\encodingdefault}{\sfdefault}{bx}{n}

\newcommand{\mup}{{\mu\text{P}}}
\newcommand{\snp}{{\textit{shrink-and-perturb}}}
\newcommand{\llm}{{LLMs}}
\newcommand{\mut}{{\mu\text{Transfer}}}

\newcommand{\szero}{{$5$M}}

\newcommand{\stwo}{{$10$M}}

\newcommand{\sfour}{{$22$M}}

\newcommand{\ssix}{{$35$M}}

\newcommand{\seight}{{$70$M}}

\newcommand{\base}{{\theta_{\text{base}}}}
\newcommand{\target}{{\theta_{\text{target}}}}
\newcommand{\targetinit}{{\theta^{0}_{\text{target}}}}
\newcommand{\fws}{{\texttt{f}_{\text{WS}}}}
\newcommand{\fwsfull}{{\texttt{f}_{\text{WS}}(\theta^{l}_{\text{base}},\hat{\theta}^{l}_{\text{target}})}}
\newcommand{\padzero}{{\texttt{pad}_0(\theta^{l}_{\text{base}},\hat{\theta}^{l}_{\text{target}})}}
\newcommand{\shrinkhp}{{\lambda_{\text{shrink}}}}
\newcommand{\stdmup}{{\sigma^{l}_{\mup}}}

\newcommand{\basem}{{\mathcal{M}_\text{base}}}
\newcommand{\targetm}{{\mathcal{M}_\text{target}}}

\newcommand{\abc}{\textit{abc}-parameterization}
\newcommand{\ws}{\textit{warmstarting}}

\usepackage{graphicx}
\usepackage{subcaption}
\usepackage{wrapfig}
\usepackage{paralist}
\usepackage{comment}

\title{Warmstarting for Scaling Language Models}
\author{%
    \begin{minipage}[t]{0.45\textwidth}
        \centering
        Neeratyoy Mallik\thanks{equal contribution,~\{mallik,~janowski\}@cs.uni-freiburg.de}
    \end{minipage} \\
    University of Freiburg\\
    \And
    \begin{minipage}[t]{0.45\textwidth}
        \centering
        Maciej Janowski$^{*}$
    \end{minipage}\\
    University of Freiburg\\
    University of Technology Nuremberg
    \And
    \begin{minipage}[t]{0.3\textwidth}
        \centering
        Johannes Hog
    \end{minipage} \\
    University of Freiburg
    \And
    \begin{minipage}[t]{0.3\textwidth}
        \centering
        Herilalaina Rakotoarison
    \end{minipage} \\
    University of Freiburg
    \And
    \begin{minipage}[t]{0.3\textwidth}
        \centering
        Aaron Klein
    \end{minipage} \\
    ScaDS.AI Leipzig
    \And
    \begin{minipage}[t]{0.45\textwidth}
        \centering
        Josif Grabocka
    \end{minipage} \\
    University of Technology Nuremberg
    \And
    \begin{minipage}[t]{0.45\textwidth}
        \centering
        Frank Hutter
    \end{minipage} \\
    ELLIS Institute Tübingen\\
    University of Freiburg
}
\begin{document}
\maketitle
\begin{abstract}

Scaling model sizes to scale performance has worked remarkably well for the current large language models paradigm.
The research and empirical findings of various scaling studies led to novel scaling results and laws that guides subsequent research.
High training costs for contemporary scales of data and models result in a lack of thorough understanding of how to tune and arrive at such training setups.
% However, prohibitively high training costs at contemporary scales of data and models result in a lack of thorough understanding of how to tune and arrive at such training setups efficiently.
One direction to ameliorate the cost of pretraining large models is to \textit{warmstart} the large-scale training from smaller models that are cheaper to tune. 
In this work, we attempt to understand if the behavior of optimal hyperparameters can be retained under warmstarting for scaling.
We explore simple operations that allow the application of theoretically motivated methods of zero-shot transfer of optimal hyperparameters using $\mut{}$.
We investigate the aspects that contribute to the speedup in convergence and the preservation of stable training dynamics  under warmstarting with $\mut{}$.
We find that shrinking smaller model weights, zero-padding, and perturbing the resulting larger model with scaled initialization from $\mup{}$ enables effective warmstarting of $\mut{}$.
\end{abstract}

\note{Keywords: warmstarting, hyperparameter optimization, language models, deep learning}

\note{TL;DR: a method to warmstart the training of a large model with optimal hyperparameter, given a tuned, trained smaller model to improve both convergence and final performance.}

\note{NM: Paper writing guide:-\\
- Consider \textit{suggest mode} for major changes. \\
- Write each sentence as a new line. \\
- Check and maximally use shortcuts/commands from \texttt{new\_commands.tex}\\
- For citations, check existence in \texttt{bib/lib.bib}, if not present, add to \texttt{bib/local.bib} \\
- Use \textbackslash{}notes\{\} for the orange block comment. \\
- Check \textit{toggle\_notes.tex} to disable notes and comments to see page count, or put notes in margin.
}

\section{Introduction}
\label{sec:intro}

Scaling of model size, dataset size and training compute together, lead to performance scaling in the large language model (\llm{}) paradigm, as shown by the recent scaling law literature~\citep{kaplan-arxiv20a, hoffmann-arxiv22a, caballero-iclr23a,hagele-icml24a,porian-icml24a}. 
Under different training setups, empirical data often fit predictable trends captured by an exponent of an assumed parametric relationship.
Recent research trends point to an attempt at understanding and finetuning the nature of scaling law studies across various problem formulations~\citep{sorscher-neurips22a, alabdulmohsin-neurips23a,wang-icml23a}.

% One commonality that persists across all such studies is the often non-overlapping choice of hyperparameter settings and documentation of how they were arrived at.
% This can be attributed to the excessively heavy costs for principled hyperparameter tuning at large scales. 
% Resuing learned weights from the tuned smaller models can be used to speed up training of larger models.

One commonality across such studies is the often non-overlapping choice of hyperparameters and sparse documentation of their selection process.
This stems from the prohibitive costs of principled hyperparameter tuning at large scales.
Reusing weights from tuned smaller models can offer a way to accelerate training of larger models.
This is commonly known as \ws{} a model's training and has been approached previously through knowledge distillation~\citep{chen-iclr16a,chen-arxiv21b}, morphisms~\citep{wei-iclr16a,elsken-iclr19a}, learned transformations~\citep{wang-iclr23a,wang-icml23a}, shrink-and-perturb~\citep{ash-neurips20a,chebykin-ecai23,shin-icml24a,samragh-arxiv24a}, heuristics~\citep{rae-arxvi21a}.
% Warmstarting a larger model's training from a trained smaller model lowers training costs by reusing learned weights~\citep{ash-neurips20a,rae-arxvi21a,deng-iccv23a,wang-iclr23a,shin-icml24a,samragh-arxiv24a}.
% In \llm{}'s sub-epoch training where data isn't repeated, successful warmstarting would amount to growing the model as data increases, as per the scaling law prescription of preference. 
% This has been approached previously in various directions 
% Various methods tackle this problem via knowledge distillation, function preserving modulation, preserving structure and rank
However, the change in optimality of hyperparameters when scaling up a model is not usually discussed alongside such sophisticated warmstarting techniques.

The growing literature on scaled parameterizations looks at the hyperparameters that directly affect a model's ability to learn features as a function of its growing model scale~\citep{yang-neurips21a,everett-icml24a,blake-icml24a,therien-arxiv24a,yang-iclr24a}.
$\mut{}$~\citep{yang-neurips21a} offers analytical scaling relations that scale the parameters for a small model: learning rate and initialization standard deviation, so that it remains optimal even for the larger model.
Our primary inquiry is to understand whether \ws{} $\mup{}$ for a large model using a \textit{tuned} smaller model would offer any improvements over vanilla-$\mup{}$. Our contributions through this work are as followed:
\begin{itemize}
    \item We identify a simple method that warmstarts $\mup{}$ runs of larger models, improving convergence speed and in certain cases final performance (RQ1 and RQ2 in Section~\ref{sec:empirical});
    % \item Empirical analysis to study the contribution of the components of the warmstarting method when transferring to different model scales;
    \item Demonstrate that warmstarting retains the $\mup{}$ training stability guarantees in practice, with respect to model scaling along specific dimensions, such as width\footnote{in this work, we look at model-width as the only scaling dimension, see Table~\ref{tab:model_scaling}} (RQ3 in Section~\ref{sec:empirical}).
    % \item (\textit{Write about Successive Warmstarting if we have something solid})
\end{itemize}

The importance of a working \ws{} method lies in its application to continually scale models with data, speeding up convergence.
% , with optimal hyperparameters.
% , and offering a way to substantially accelerate the tuning process.
We believe that enabling warmstarted-$\mup{}$ can also substantially accelerate the tuning of non-$\mu$-parameterizable hyperparameters for large models.
% {\color{red}(see, Section~\ref{})}.

% \note{Aaron: I think we should better motivate the high level goal of this line of research. What is bigger problem we try to address? For example, we could highlight that in the end we expect this to pave the way to substantially accelerate the HPO process.\\
% NM: Edited the line a bit}
In the next section, we briefly highlight the method (Section~\ref{sec:method}) we find to be the most simple and yet in line with scaled parameterizations, allowing us to leverage algorithms such as $\mup{}$.
We then show our empirical findings that validate our approach (Section~\ref{sec:empirical}).
Appendix~\ref{app:bg-related} onwards contain various supporting information and details.

\section{Method: Warmed-$\mup{}$}
\label{sec:method}

In this section, we formalize and formulate warmstarting in the context of scaled parameterizations. We refer the readers to Appendix~\ref{app:bg-related} for more on background concepts and related work.

Given a trained \textit{base} model $\basem{}$ with tuned hyperparameters (e.g. learning rate) and a \textit{target} model $\targetm{}$ for transferring the hyperparameters, we define the \ws{} operation as the layer-wise initialization of $\targetm{}$ using $\basem{}$. In this work, we initialize each layer of $\targetm{}$ using a combination of a shrunk version of $\basem{}$ weights with the standard $\mup{}$ initialization. For a layer $l$ in $\targetm{}$, let $\theta_{\text{target}}^{l}\in\mathbb{R}^{p \times q}$ be its weight and $\theta_{\text{base}}^{l}\in\mathbb{R}^{m \times n}$ be the corresponding weight from $\basem{}$ (where $m \le p$ and $n \le q$). Formally, the warmstarting operation is given by:

\begin{equation}
\theta_{\text{target}}^{l}~=~\shrinkhp{} \cdot \texttt{Pad}_0{}(\theta_{\text{base}}^{l}, p, q) + \mathcal{N}(0,~\stdmup{}^{2}), 
\label{eq:warm-snp}
\end{equation}
where $\stdmup{}$ is the recommended per-layer standard deviation by~\citet{yang-neurips21a} for initialization, and $\texttt{Pad}_0$ is a function that expands the \textit{base} model weight to the \textit{target} shape ($p\times q$) with zero padding. Intuitively, the initialization term of Equation~\ref{eq:warm-snp} can be replaced with any scaling parameterization technique, while the first term can represent any transformation of the base weight matrices onto the target shape. 
Additionally, setting $\shrinkhp{}=0$ in Equation~\ref{eq:warm-snp} recovers vanilla-$\mup{}$.

We choose zero-padding as the simplest method to study for scaling the model size with $\mup{}$ and avoid more sophisticated heuristics~\citep{rae-arxvi21a}, learned transformations~\citep{wang-iclr23a} or distillations~\citep{chen-iclr16a}.
%The $\stdmup{}$ is the layer-wise scalar as obtained from the the $\sigma_{\text{base}}$ used to train $\basem{}$, scaled as per $\mup{}$ prescriptions~\citep{yang-neurips21a}.
% \note{Aaron: We should motivate why scaling down the base model weights is important. Also we should explain how this maps to the definition of $\fws{}$.\\
% NM: Tried to do so.
% }
In order to retain $\mup{}$'s guarantees of training stability, $\shrinkhp{} \in [0,~1]$ is multiplied to all the transformed weights.
The $\mup{}$ initialization acts as a \textit{perturbation} matrix for the \textit{base} model weights, transformed and casted to the \textit{target} shape.
We choose a fixed value of $\shrinkhp{}=0.4$ in our experiments (Section~\ref{sec:empirical}) as per \snp{}-based \ws{} literature~\citep{ash-neurips20a,zaidi-neurips22a,chebykin-ecai23} from the continual learning setting without model size growth.
We find empirically that this recommendation interestingly holds and is crucial for warmstarted-$\mup$ (see, Section~\ref{sec:empirical}, Appendix~\ref{app:exp-train-dyna}).

From a different perspective, one can also see the \ws{} operation as equivalent to \abc{}, which defines the family of scaled parameterization methods~\citep{yang-icml21a,blake-icml24a,everett-icml24a} that derive fixed scaling rules for hyperparameter transfer \citep{yang-neurips21a}.
Scaled parameterization refers to modifying the initialization ($\mathcal{B}_w$), weight parameter scaling ($\mathcal{A}_w$) and learning rate ($\mathcal{C}_w$) as a function of model scales (see, Appendix~\ref{app:bg-related}).
Designing Equation~\ref{eq:warm-snp} in a lean manner opens up the direction of phrasing $\shrinkhp{}$ as $\mathcal{A}_w$, while the overall scaled initialization can now be, $\sim \mathcal{N}(w_{0}', \mathcal{B}_{w}^{2})$, where $w_{0}'~=~\shrinkhp{} \cdot \texttt{Pad}_0{}(\theta_{\text{base}}^{l}, p, q)$.
The next section shows empirically the gains provided by such simple adjustments and why studying and understanding warmstarted-$\mup{}$ more is relevant.

\section{Empirical evaluation}
\label{sec:empirical}

In this section, we report our empirical findings, with the intention of emphasizing the underline intuition behind the proposed approach (Section~\ref{sec:method}), in both convergence speed and training stability.

% \note{Aaron: We should add a paragraph to describe our empirical setup: which model, architecture, dataset, etc }

\paragraph{Experimental Setup}: We train a decoder-only GPT2 model~\citep{radford-openaiblog19a} as per the GPT-NeoX implementation~\citep{gpt-neox23} provided by LitGPT~\citep{litgpt23} on the SlimPajama~\citep{slimpajama23}'s $6$B version\footnote{\url{https://huggingface.co/datasets/DKYoon/SlimPajama-6B}} dataset.
We set the weight decay to $0$ for Adam~\citep{kingma-iclr15a} to avoid interaction effects and confounding factors~\citep{lingle-arxiv24a,wang-arxiv24a}.
Fixed learning rate (LR) schedules were used following~\citet{hagele-icml24a}, simplifying checkpoint studies with no LR management overheads. 
% We replicated~\citet{hagele-icml24a}'s findings with a constant LR schedule in our setup.
We disable learning rate warmup to aid comparative analysis of \textit{if} warmstarting provides gains.
All models are trained for $20$ tokens/parameter, similar to~\citet{hoffmann-arxiv22a}, keeping sub-epoch training, that is, no micro batch of the dataset is ever repeated. 
Our FLOPs calculation is also based on~\citet{hoffmann-arxiv22a}.
Validation loss is measured on a fixed held-out split of the dataset, consistent across all runs, with $3$ different seeds per run, and Gaussian smoothing is used for loss comparison across runs.
For \ws{} runs, we do not repeat tokens seen by the \textit{base} model. 
A \ws{} training starts with a fresh optimizer state, and follows standard $\mup{}$ training procedures. All trainings used a single NVIDIA RTX 2080 GPU.

% \note{
% NM: Most crucial aspects to include: architecture, dataset, training budget choice, parameter count choice, optimizer choice (0 WD Adam), and LR schedule choice (constant), vocab size, tokenizer, etc.

% }

% \begin{itemize}
%     \item Only the high-level detail required to read our subsequent plots (decoder-only GPTNeoX style or whatever, no grouped attention, only width scaling (point to table in Appendix ({\color{red}TODO}), Slimpajama, chinchilla budget, constant LR, no weight decay Adam (point to HP table in Appendix ({\color{red}TODO})).
%     \item For everything else, point to Appendix~\ref{app:exp-setup}.
%     \item Mention validation loss over fixed held-out set. 
%     \item Importantly, we do not repeat tokens seen by a smaller model when warmstarting a bigger one from it.
%     \item Might add a line here on our choice of x-axis and their equivalence (could point to Appendix for a more detailed version of Johannes' calculations)
% \end{itemize}

% \subsection{Hypothesis 1: Does warmstarting improve a vanilla-$\mup{}$ run?}
\paragraph{RQ 1: Does warmstarting improve a vanilla-$\mup{}$ run?}
% \label{sec:warm-better-mup}

\begin{figure}[htbp]
\centering
\begin{tabular}{c}
    \includegraphics[width=0.9\columnwidth]{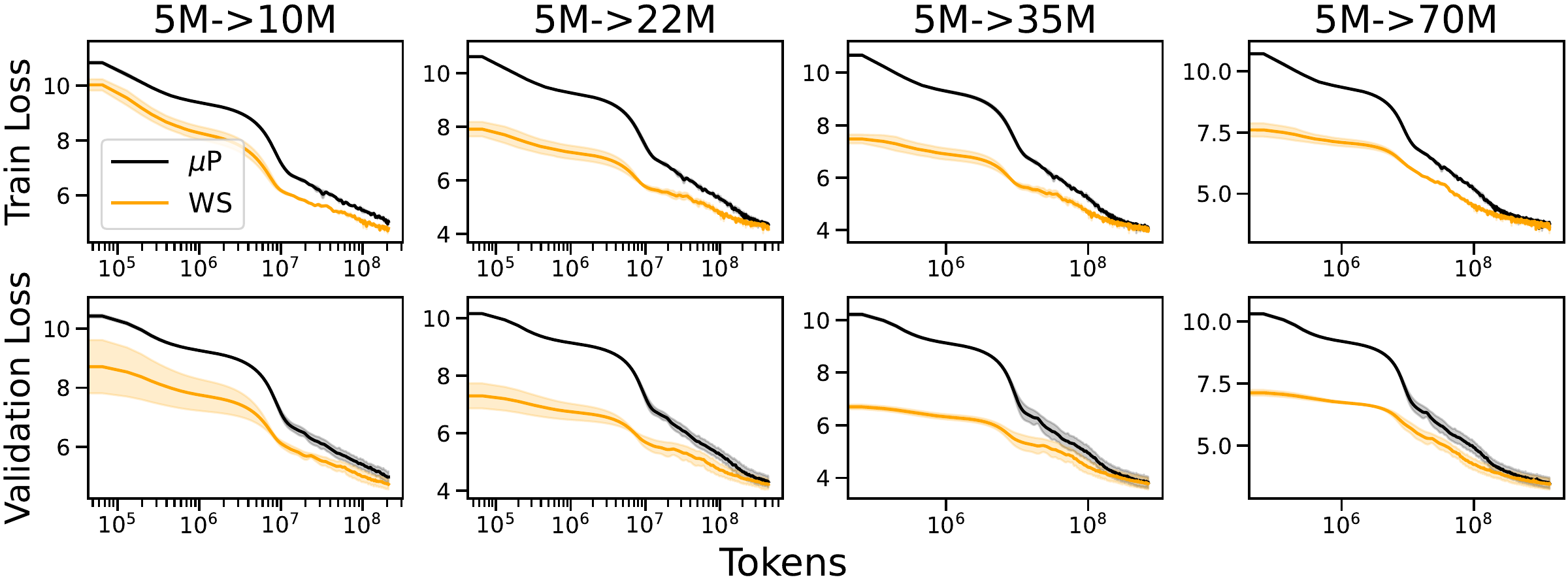}
\end{tabular}
\caption{
Transferring the best found learning rate at the base scale of \szero{} using $\mup{}$.
For \ws{} (WS) run, the model weights of the optimal \szero{} model is used to initialize the \textit{target} model's training.
Warmstarting appears to always improve $\mup{}$ convergence rates.
}
\label{fig:warm-better-mup}
\end{figure}

The $\mup{}$ recipe recommends the following broadly for $\mut{}$:
\begin{inparaenum}[i)]
    \item Perform a grid search over hyperparameters at a small model scale;
    \item Transfer the optimal hyperparameters at this scale, to a larger model scale using $\mup{}$ rules.
\end{inparaenum}
The exact choice of \textit{base} and \textit{target} scales, choice of hyperparameters for grid search, if the base scale itself should be a hyperparameter, etc. are all actively pursued investigations in the community~\citep{everett-icml24a, blake-icml24a, qiu-icml24a, lingle-arxiv24a}.
Given our experimental setup of fixed model scales across width scaling (refer, Table~\ref{tab:model_scaling}), we choose $3$ possible \textit{base} scales: \szero{}, \stwo{} and \sfour{}.
Figure~\ref{fig:warm-better-mup} shows $\mut{}$ from the smallest base scale with and without warm-starting.
Figure~\ref{fig:warm-longer} shows the same result as Figure~\ref{fig:warm-better-mup} but under longer compute or more tokens than recommended by~\citet{hoffmann-arxiv22a}.
Figure~\ref{fig:warm-mup-full-set} reports \ws{} from other base scales, scaled up to a 120M model and Figure~\ref{fig:warm-mup-full-set-spike} shows learning curves that include the base model training.

Warmstarting (WS), even across much larger model sizes yield a gain in initial performance, while either matching or improving the vanilla-$\mup{}$ performance under equivalent compute (as per~\citet{hoffmann-arxiv22a}).
Here, we compare the runs over the training compute as required by the target model.
Both $\mup{}$ and warmstarting runs require a grid search from the \textit{base} scale (\szero{}) and hence are expected to share similar compute, and thus can be omitted from this analysis (Figure~\ref{fig:warm-better-mup}).

\paragraph{RQ 2: How does the scale difference affect the quality of warmstarting?}

\begin{figure}[htbp]
    \centering
    \begin{tabular}{c}
    \includegraphics[width=0.95\columnwidth]{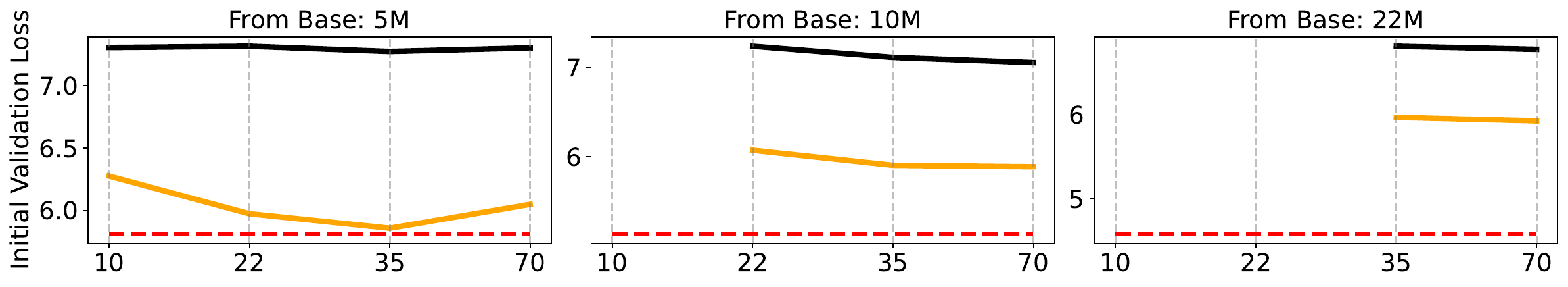} \\
    \includegraphics[width=0.95\columnwidth]{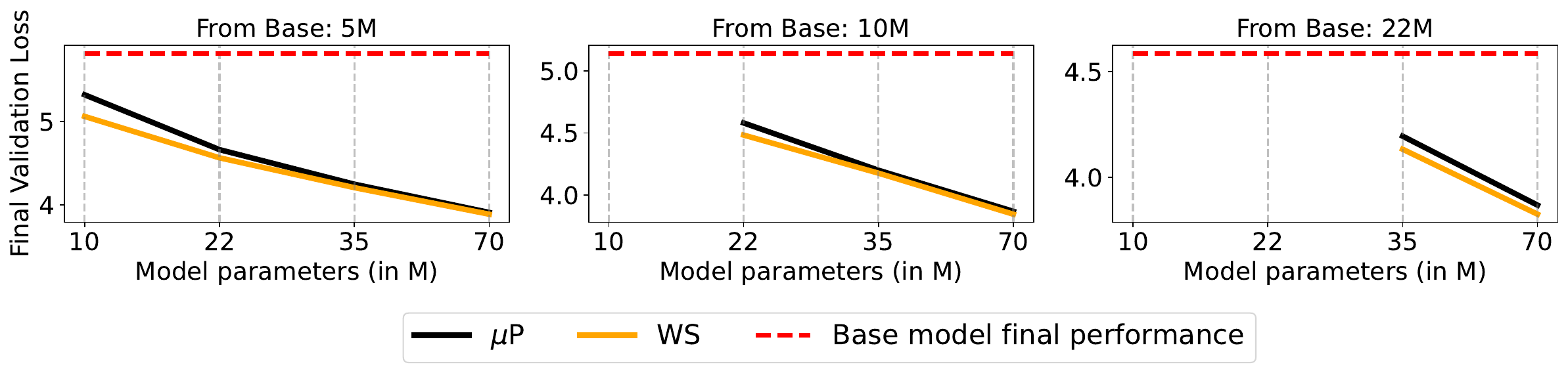}
    \end{tabular}
    \caption{
    % {\color{red}TODO.}
    Comparing losses across model scales. (\textit{Left to right}): given a larger \textit{base} model, transfer to higher model scales; (\textit{Top}): Shows the initial validation loss of the warmstarted model vs. vanilla-$\mup{}$, where \ws{} always leads to improved initial loss; (\textit{Bottom}): Shows the final validation loss of the warmstarted model vs. vanilla-$\mup{}$ run, which achieves better or equivalent loss.
    % Here, the validation loss is computed using Gaussian smoothing over 25 steps. Total model parameters count account for embedding parameters.
    }
    \label{fig:ws-mup-scaling}
\end{figure}
% \note{NM: Can we do plot with seeds of this? \\
% NM: I tried, but with seeds there is no conclusive evidence at all for the initial loss (or I have a plotting bug (though unlikely since the final loss plot lines up))
% }

Given that the warmstarting operation %$\fws{}$ 
(Section~\ref{sec:method}) chosen for our experiments pads the new connections to the \textit{base} model with zeros to scale to the \textit{target} model, it can be expected that the quality of warmstarting will be more pronounced when a \textit{large enough} model is grown not \textit{too large}.
% proportional to the model scale difference between \textit{target} and \textit{base}.
In Figure~\ref{fig:ws-mup-scaling} we chart the initial and final validation loss comparison with $\mut{}$ and warmstarting under width-scaling.
We observe that warmstarting offers a much-improved initialization that accelerates loss convergence.
It is also evident that the loss obtained by the \textit{base} model is not transferred.
This gap manifests as \textit{spikes} if learning curves are concatenated over the \textit{base} and warmstarted-\textit{target} runs (see Figure~\ref{fig:warm-mup-full-set-spike}).
Figure~\ref{fig:ws-mup-scaling} also highlights that the amount of data already seen by the base model, and the loss it can achieve given its scale, skews such an analysis as the resulting warmstarted model is likely to have a larger spike in loss.
Given that we do not repeat the tokens seen by the \textit{base} model when training the warmstarted \textit{target} model, a larger spike would require more updates and thereby tokens to recover.
We believe that though the \ws{} method shown here is adequate in providing speed ups, it is suboptimal in not providing consistent \textit{improvement} in final loss, despite seeing more tokens in principle.
Despite the obvious gains \ws{} can bring with $\mup{}$, there are clearly much more design choices to study and consider for consistent, efficient warmstarting.

\paragraph{RQ 3: Does warmstarting with $\mup{}$ retain training stability guarantees?}

\begin{figure}[htbp]
    \centering
    \includegraphics[width=\columnwidth]{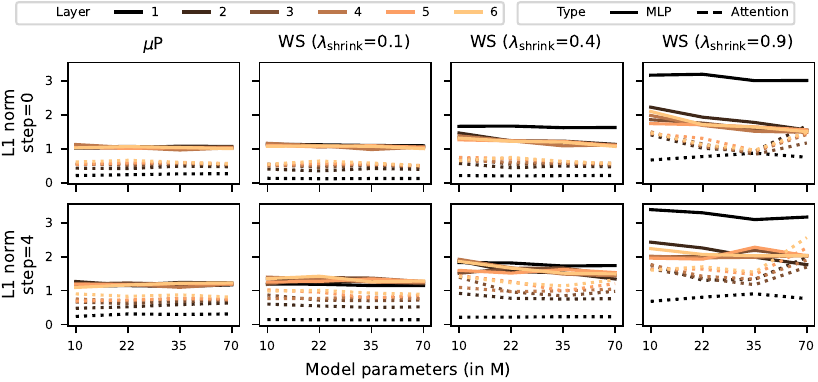}
    \caption{L1 norm of the layers activation across scales. Any warmstarted $\mup{}$, having $\shrinkhp{}~\le~0.6$, behaves well in scale as in $\mup{}$ (detailed results in Figure~\ref{fig:coord_checks_full} of Appendix~\ref{app:exp}).}
    \label{fig:coord_checks}
\end{figure}

We empirically investigate the impact of the proposed \ws{} technique on $\mup{}$ training behavior~\citep{yang-neurips21a}. 
% Retaining $\mup{}$ is indeed essential to ensure optimal hyperparameter transferability and training stability across scales. 
% To this end, 
We conduct coordinate checks as per the $\mup{}$ library\footnote{\url{https://github.com/microsoft/mup?tab=readme-ov-file\%23coord-check}} to verify the consistency of the $L_1$-norm of layer activations across width-scaling. 
Within the same experimental setting, we also ablate the effect of the shrinking value, the main hyperparameter of our approach.

Figure \ref{fig:coord_checks} shows L1-norm of activations for warmstarted $\mup{}$ with $\shrinkhp~\le~0.6$ trend similar to vanilla $\mup{}$ across the considered width-scaling. 
However, we observe instability for  $\shrinkhp~>~0.6$ (Figure~\ref{fig:coord_checks_full},~\ref{fig:warm-shrink-check}). 
This result also backs the $\shrinkhp = 0.4$ value reported for \snp{} in the literature~\citep{ash-neurips20a,zaidi-neurips22a,chebykin-ecai23}.
Additionally, we monitor the L1 norm of layer activations during the entire training in Figure~\ref{fig:warm-mup-l1-stable}. 
Warmstarted $\mup{}$ maintains activation values within a comparable range and trends consistent with vanilla-$\mup{}$, suggesting that our demonstrated warmstarting method does not introduce instability to $\mup{}$.

%\note{better motivation for shrink and perturb via coord check plots, instead of the loss ones, not sure if we want that in the appendix either}

\begin{comment}

\begin{itemize}
    \item Show activation L1 plots for different components: WS vs $\mup{}$: scale0 \textrightarrow scale8, scale4 \textrightarrow scale6
    \item Show L1 and L2 of weights (can we show mean using per layer): WS vs $\mup{}$: scale0 \textrightarrow scale8, scale4 \textrightarrow scale6
    \item Compare other more naive baselines such as \textit{zeros} and \textit{zeros\_mup} to highlight that \textit{both} shrinking and \mup-perturb are required (and \textit{zeros\_with\_shrinking} and \textit{zeros\_with\_perturbation}) - Figure~\ref{fig:fig3}.
    \item Point to the ablation of s0-s8 shrink \ref{fig:shrink} and perturb \ref{fig:perturb} (run for s4-s6 as well)
\end{itemize}

\begin{figure}[htbp]
\centering
\begin{tabular}{cc}
    \includegraphics[width=0.5\columnwidth]{neurips2024_ws/figures/h3/fig3_0_8.pdf} & \includegraphics[width=0.5\columnwidth]{neurips2024_ws/figures/h3/fig3_2_4.pdf} \\
\end{tabular}
\caption{}
\label{fig:fig5}
\end{figure}
\end{comment}

% \begin{figure}
%     \centering
%     \includegraphics[width=0.5\linewidth]{neurips2024_ws/figures/screenshot.png}
%     \caption{}
%     \label{fig:screenshot}
% \end{figure}

\section{Conclusion}

In this work, we explore the feasibility of enabling continued pretraining of language models as more data is provided and model size is scaled up.
We demonstrate that a simple technique, even when warmstarting from a much smaller model, is adequate in improving convergence speed and potentially final performance (Figures~\ref{fig:warm-better-mup},~\ref{fig:warm-mup-full-set}).
Our focus was to identify a warmstarting method that is simple enough to be implemented alongside $\mup{}$ (Section~\ref{sec:method}) while not being in conflict with $\mut{}$'s expectations of stable training in practice (Figures~\ref{fig:coord_checks},~\ref{fig:coord_checks_full},~\ref{fig:warm-mup-l1-stable},~\ref{fig:warm-shrink-check}).
% We demonstrate exactly so under a fixed training setup.

\paragraph{Limitations.} We note that the empirical study shown here has an outcome of interest in that simple shrinking and perturbation with $\mup{}$ works already.
However, our setting (Section~\ref{sec:empirical}) is carefully chosen to minimize effects induced by various choices of training setups, in order to understand the effects of \ws{} with $\mup{}$.
For more general practicality, more ablations and experiments must be performed over different learning rate schedules, weight decay, activations, and for much larger model sizes and context windows.
More recent works on scaled parameterizations~\citep{everett-icml24a,blake-icml24a} which  study and improve over $\mup{}$ recommendations should also be tested with our modular approach to further assert the potential of \ws{} under \abc{}.

\paragraph{Future directions.}
More diverse experiments aside, exploring what makes \ws{} work, as we demonstrate here, is important for a lean, general, efficient method that truly speeds up pretraining runs.
Figures~\ref{fig:coord_checks_full} and~\ref{fig:warm-shrink-check} show the effect of different shrinking factors in maintaining training stability of optimal hyperparameters, while Figure~\ref{fig:ws-mup-scaling} shows that choice of scales directly affect gains provided by \textit{warmstarting}.
Exploiting the definition of Equation~\ref{eq:warm-snp}, we would like to explore the role of \textit{layer-wise} shrinking of base model weights and investigate theoretical relations with \abc{}.
Exploiting structures, ranks, and representations learned by the base model can also lead to better warmstarting and improvement over simple zero padding~\citep{rae-arxvi21a,wang-icml23a,qiu-icml24a,wei-arxiv24a}.
Moreover, warmstarting using our setup here leads to the consumption of more tokens, when comparing against the vanilla-$\mup{}$ training of the \textit{target} model.
Exploring if models can be warmstarted and scaled up progressively will be a clear direction to pursue (see, Appendix~\ref{sec:successive-warms}).
Such a paradigm of training can massively lower hyperparameter tuning costs overall and also potentially change the practice of scaling models up, with tuned hyperparameters.

\newpage

% \begin{acknowledgements}

\paragraph{Acknowledgements.}
\textbf{FH} acknowledges the financial support of the Hector Foundation.
\textbf{NM}, \textbf{JH}, \textbf{HR}, \textbf{FH} acknowledge funding by 
the state of Baden-W\"{u}rttemberg through bwHPC, the German Research Foundation (DFG) through grant numbers INST 39/963-1 FUGG and 417962828, and the European Union (via ERC Consolidator Grant Deep Learning 2.0, grant no.~101045765), TAILOR, a project funded by EU Horizon 2020 research and innovation programme under GA No 952215. 
Views and opinions expressed are however those of the author(s) only and do not necessarily reflect those of the European Union or the European Research Council. 
Neither the European Union nor the granting authority can be held responsible for them.
\textbf{JH} and \textbf{FH} acknowledge funding by the Deutsche Forschungsgemeinschaft (DFG, German Research Foundation) under SFB 1597 (SmallData), grant number 499552394.
\textbf{MJ} and \textbf{JG} acknowledge funding by The Carl Zeiss Foundation through the research network "Responsive and Scalable Learning for Robots Assisting Humans" (ReScaLe) of the University of Freiburg.
\textbf{AK} acknowledges the financial support by the Federal Ministry of Education and Research of Germany and by Sächsische Staatsministerium für Wissenschaft, Kultur und Tourismus in the programme Center of Excellence for AI-research ``Center for Scalable Data Analytics and Artificial Intelligence Dresden/Leipzig'', project identification number: ScaDS.AI.
\begin{center}\includegraphics[width=0.3\textwidth]{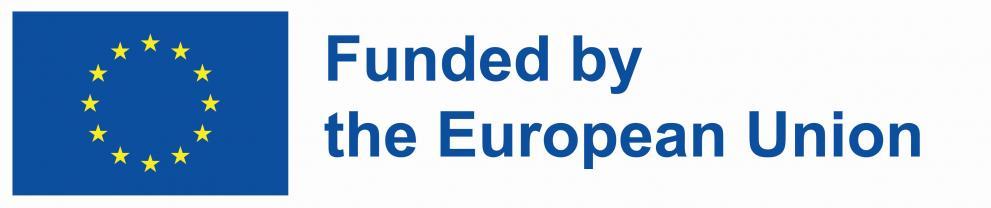}\end{center}

% \end{acknowledgements}

% \begin{figure}
%     \centering
%     \includegraphics[width=0.9\linewidth]{neurips2024_ws/figures/eigenvalues_s0_s6.pdf}
%     \caption{Scale0->6}
%     \label{fig:eigen}
% \end{figure}

% \subsection{Limitations}

% \ldots

% \subsection{Future work}

% PMS lays the groundwork for integrating model size as a tunable parameter in grey-box hyperparameter optimization (HPO) methods, such as Hyperband. 
% Traditionally constrained by the computational overhead of training from scratch, our approach mitigates this through efficient warmstarting from smaller models. 
% This allows for a broader exploration across multiple fidelity dimensions, including both epochs and model size. 
% Future research will explore this integrated approach's potential to enhance model performance and optimization efficiency, expanding its applicability across various architectures and datasets.

%  If you'd like to, you may include a section for author contributions as is done


% \newpage

% \bibliographystyle{neurips2024_ws/neurips_2024}
\bibliographystyle{unsrtnat}
\bibliography{bib/lib,bib/local,bib/proc,bib/strings}

\newpage

\appendix

\section{Background and Related Work}
\label{app:bg-related}

\paragraph{Scaled parameterizations} 
refers to a set of rules that determine how certain parameters can be scaled with respect to one or more scaling dimensions~\citep{everett-icml24a}.
Such prescriptions attempt to ensure stable feature learning under stable optimal hyperparameters to ensure maximal feature learning in the infinte-width limit~\citep{yang-icml21a}.
\abc{}s~\citep{yang-icml21a} is a formulation that subsumes such parameterizations and broadly follows the assumption where model weights are so defined,

% \begin{align}
% \begin{split}
%     w_0 \sim \mathcal{N}(0, \boldsymbol{\mathcal{B}^{2}_{w}}) \\
%     W_t = \boldsymbol{\mathcal{A}_w} \cdot w_t \\
%     w_{t+1} = w_{t} + \boldsymbol{C_w} \cdot \Phi_t(\nabla L_0,\ldots, \nabla L_t)
% \end{split}
% \end{align}

\begin{align}
    w_0 \sim \mathcal{N}(0, \boldsymbol{\mathcal{B}^{2}_{w}}),
\end{align}
\begin{align}
    W_t = \boldsymbol{\mathcal{A}_w} \cdot w_t,
\end{align}
\begin{align}
    w_{t+1} = w_{t} + \boldsymbol{\mathcal{C}_w} \cdot \Phi_t(\nabla L_0,\ldots, \nabla L_t),
\end{align}
where $t$ defines the training step and $\Phi_t(\nabla L_0,\ldots, \nabla L_t)$ is the gradient-based weight update step~\citep{blake-icml24a}.
$\mathcal{A}_w,~\mathcal{B},~\mathcal{C}_w$ are scalars that change with model width and determine the required scaling of parameterized entities.
The specific prescription of these scalars is governed by the method and its assumptions which lead to the analytical rules of scaling.
We refer the reader to Tables 1, 2, 3 from~\citet{yang-neurips21a} for a succinct summary of the scaling rules for the scalars $\mathcal{A}_w,~\mathcal{B},~\mathcal{C}_w$, which realizes $\mu$Parameterization or $\mup{}$.
This class of methods is seeing growing interest in the community as model sizes increase and prior knowledge of good hyperparameters are not available on such scales~\citep{yang-iclr24a,everett-icml24a,blake-icml24a,therien-arxiv24a,qiu-icml24a,lingle-arxiv24a}.
Recent studies find improvements in the $\mup{}$ recommendation and scenarios or setups where $\mup{}$ can be bettered.
Our work is along these lines where we aim to show that \ws{} a larger model training from a smaller model is one more such scenario where vanilla-$\mup{}$ can be improved upon.

\paragraph{Shrink-and-perturb} is a technique that showed that models of the same size can be warmstarted from a previous training run successfully such that it converges faster without hurting generalization~\citep{ash-neurips20a}. 
Under \snp{}, every learnable parameter $w^{i}_{t}$ is initialized as,
$w^{i}_{t} \leftarrow \lambda \theta^{i}_{t-1} + p_t$, where 
$p_t \sim \mathcal{N}(0,~\sigma^2)$ and $0 < \lambda < 1$.
~\citet{ash-neurips20a} posits that the shrinking of the weights act as a regularization but not similar to weight decay.
In effect, shrinking weights can act as increasing the entropy of the output distribution and preserving the relative activation at each layer.
While perturbation seems to have an effect on balancing gradient contribution from each learned parameter.
Crucially, this version of \snp{} was specifically shown for continual learning under stationary data distributions~\citep{chebykin-ecai23,shin-icml24a}.
However, to our knowledge, no such \snp{} warmstarting method directly talks about warmstarting across model scales.
% The literature on warmstarting models across model sizes is rich and contains various different flavours of approaches including, but not limited to, initializations~\citep{zaidi-neurips21a}, morphisms~\citep{wei-icml16a}, distillation and transfer~\citep{chen-iclr16a,chen-arxiv21b,wang-icml23a}, learned transformations~\citep{deng-iccv23a,wang-iclr23a}, heuristics~\citep{rae-arxvi21a}.
% However, to our knowledge, none of these methods discuss a principle strategy for setting hyperparameters when growing the model.
% Moreover, each method is a well-designed, sophisticated method and often make assumptions.
% Whereas our target is to bring forth a theoretically motivated modular approach that, given any model checkpoint with its tuned hyperparameters, can scale this model along certain scaling dimensions, initialize by warmstarting with the smaller model weights, and train using suitably scaled hyperparameters.
% ~\citep{samragh-arxiv24a,yao-iclr24a,du-neurips24a}.

\paragraph{Model growth} 
literature, or warmstarting literature across model sizes is rich and encompasses multiple methodological paradigms, including initialization techniques~\citep{zaidi-neurips21a}, network morphisms~\citep{wei-icml16a}, knowledge transfer and distillation approaches~\citep{chen-iclr16a,chen-arxiv21b,wang-icml23a}, learned transformations~\citep{deng-iccv23a,wang-iclr23a}, and empirical heuristics~\citep{rae-arxvi21a}. 
More recent works show that increasing transformer model size can improve pretraining efficiency and potentially the compute-loss scaling coefficient~\citep{yao-iclr24a,du-neurips24a}.
Each such method represents a well-designed and sophisticated approach that, in principle, achieves the required warmstarting effect. 
These approaches conflict with our requirement for a simple and practical implementation, require changes to training routines, loss functions, and usually keep the hyperparameters fixed across scales.
~\citet{samragh-arxiv24a}, a concurrent work, echoes our requirements for a warmstarting method and proposes a \snp{}-like method for growing a language model, and is the closest work to us, altering only the initialization method given a smaller \textit{base} model checkpoint.
Our method and approach differ from all approaches mentioned above since we study and demonstrate that it is possible to warmstart \textit{while} suitably scaling hyperparameters.

Our work aims to develop a theoretically motivated modular framework that can, given any model checkpoint with its tuned hyperparameters, systematically scale this model along arbitrary dimensions, initialize through principled warmstarting from smaller model weights, and train using theoretically derived hyperparameter scaling rules.

\section{Experiments}
\label{app:exp}

This section covers additional details on the experimental setup and more supporting results.

\paragraph{Model scales.} Table~\ref{tab:model_scaling} covers the model scales reported in the experiments in this work.

\begin{table}[h!]
\centering
\begin{tabular}{cccccc}
\hline
\textbf{n\_layers} & \textbf{d\_model} & \textbf{n\_head} & \textbf{head\_size} & \textbf{\# params [M] } \\ 
\hline
$6$ & $48$ & $2$ & $24$ & $5$ \\ 
$6$ & $96$ & $6$ & $16$ &  $10$ \\ 
$6$ & $192$ & $8$ & $24$ &  $22$ \\ 
$6$ & $288$ & $12$ & $24$ &  $35$ \\ 
$6$ & $512$ & $16$ & $32$ &  $70$ \\ 
% $6$ & $768$ & $24$ & $32$ &  $120$ \\ 
% $6$ & $1024$ & $32$ & $32$ &  $179$ \\ 
\hline
\end{tabular}
\caption{
Overview of model scaling parameters with the number of layers fixed at 6 across all configurations. 
The \(d\_model\) parameter is defined as the product of the embedding size and the number of attention heads. 
The \textit{block size} is set to $1024$. 
The effective batch size is as determined by the grid search on a base scale used for $\mut{}$.
% The number of parameters is calculated using Kaplan's formula.
% \note{NM: should report \textit{effective batch size} or should remove the batch size from here: here it is the max micro batch size for a specific hardware}
}
\label{tab:model_scaling}
\end{table}

% \note{NM: Can we get Table~\ref{tab:model_scaling} to be something like Table 1 in \url{https://arxiv.org/abs/2407.09835}}

\paragraph{Grid search results for \textit{optimal} hyperparameters.}

For the chosen $3$ base scales, we perform a grid search over the learning rate (LR) and batch size. 
The grid search results are summarized below:
\begin{itemize}
    \item \szero{}: $\texttt{learning rate}=0.03$; $\texttt{batch size}=64$
    \item \stwo{}: $\texttt{learning rate}=0.01$; $\texttt{batch size}=64$
    \item \sfour{}: $\texttt{learning rate}=0.003$; $\texttt{batch size}=256$
\end{itemize}
When performing $\mut{}$ in our work, the batch size found for the base scale is kept constant while LR is scaled as per $\mup{}$.

\subsection{Additional results}

\begin{figure}[htbp]
    \centering
    \includegraphics[width=0.95\linewidth]{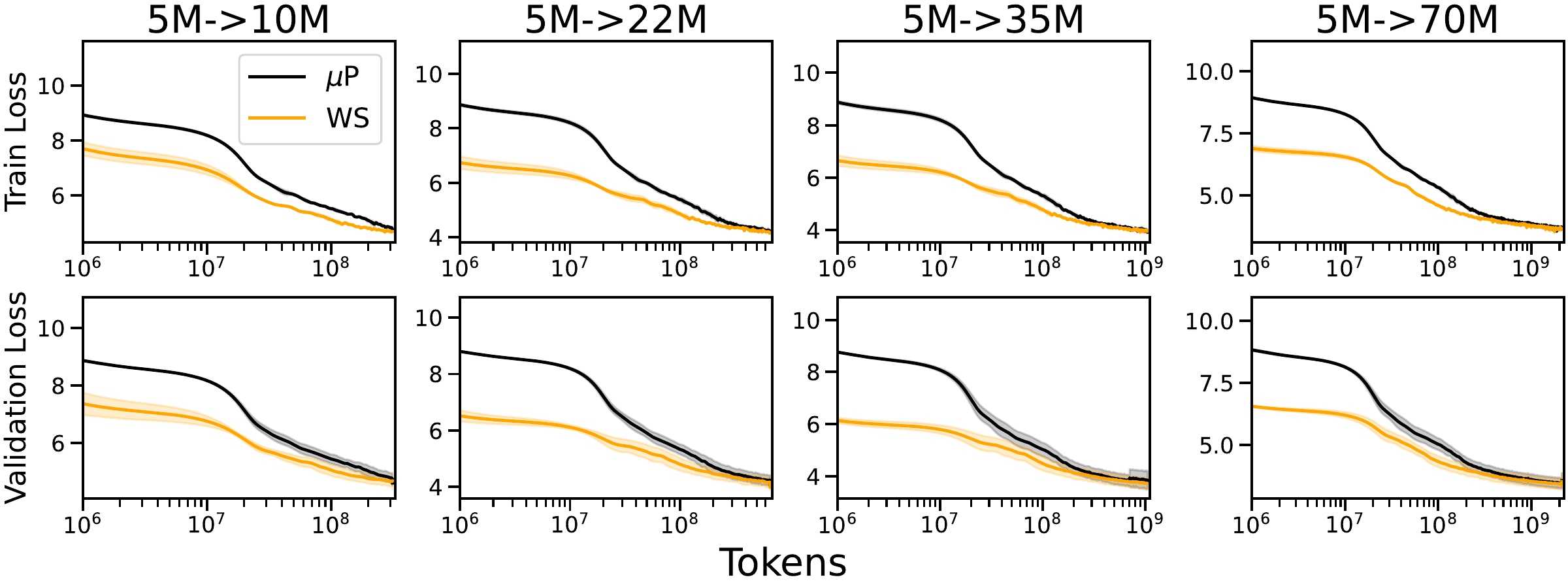}
    \caption{Models in Figure~\ref{fig:warm-better-mup} trained for more tokens and thus compute. Here, we train each model for 30 tokens/parameter instead of the 20 recommended by~\citet{hoffmann-arxiv22a}.}
    \label{fig:warm-longer}
\end{figure}

\begin{figure}[htbp]
    \centering
    \begin{tabular}{c}
        \includegraphics[width=0.88\columnwidth]{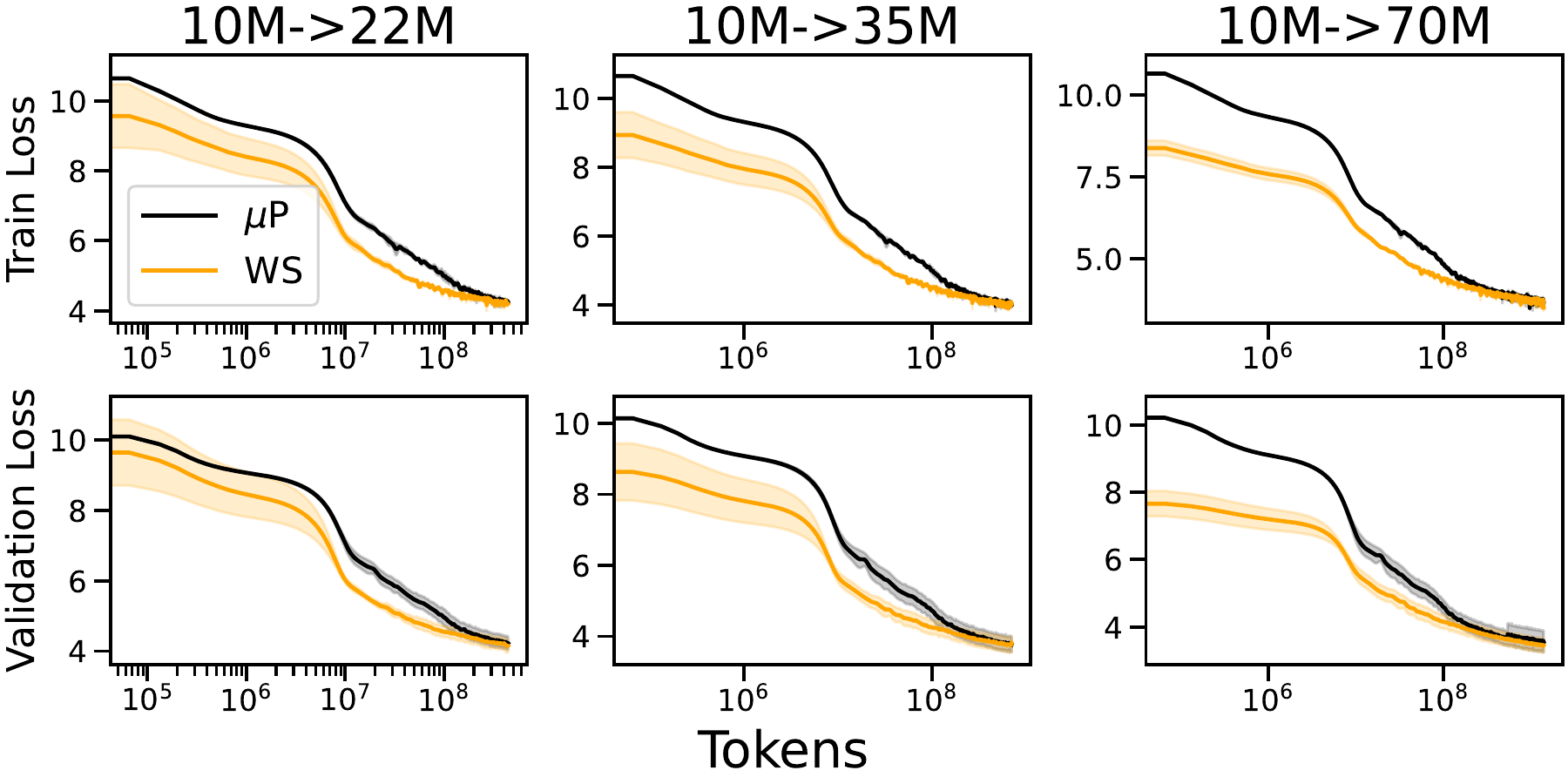} \\
        \includegraphics[width=0.88\columnwidth]{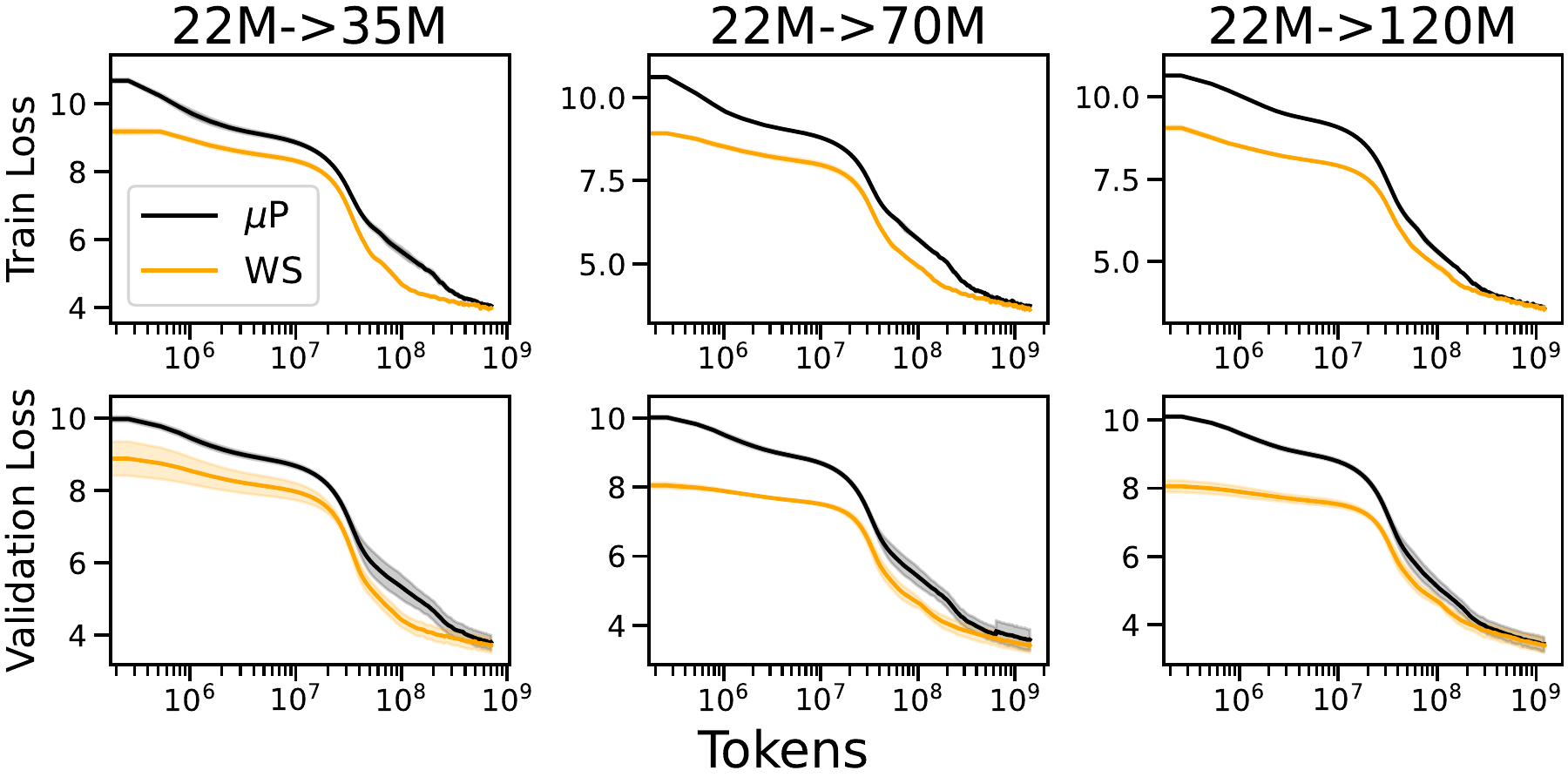} \\
    \end{tabular}
    \caption{
    Transferring the best found learning rate at the base scale of \stwo{} (\textit{top set of $2\times3$}) and \sfour{} (\textit{bottom set of $2\times3$}) using $\mup{}$.
    For \ws{} (WS) run, the model weights of the optimal \textit{base} model is used to initialize the \textit{target} model's training.
    Warmstarting improves $\mup{}$ convergence, though the quality speedup and gains depend heavily on the choice of base and target scales.
    }
    \label{fig:warm-mup-full-set}
\end{figure}

\begin{figure}[htbp]
    \centering
    \begin{tabular}{c}
        \includegraphics[width=0.9\columnwidth]{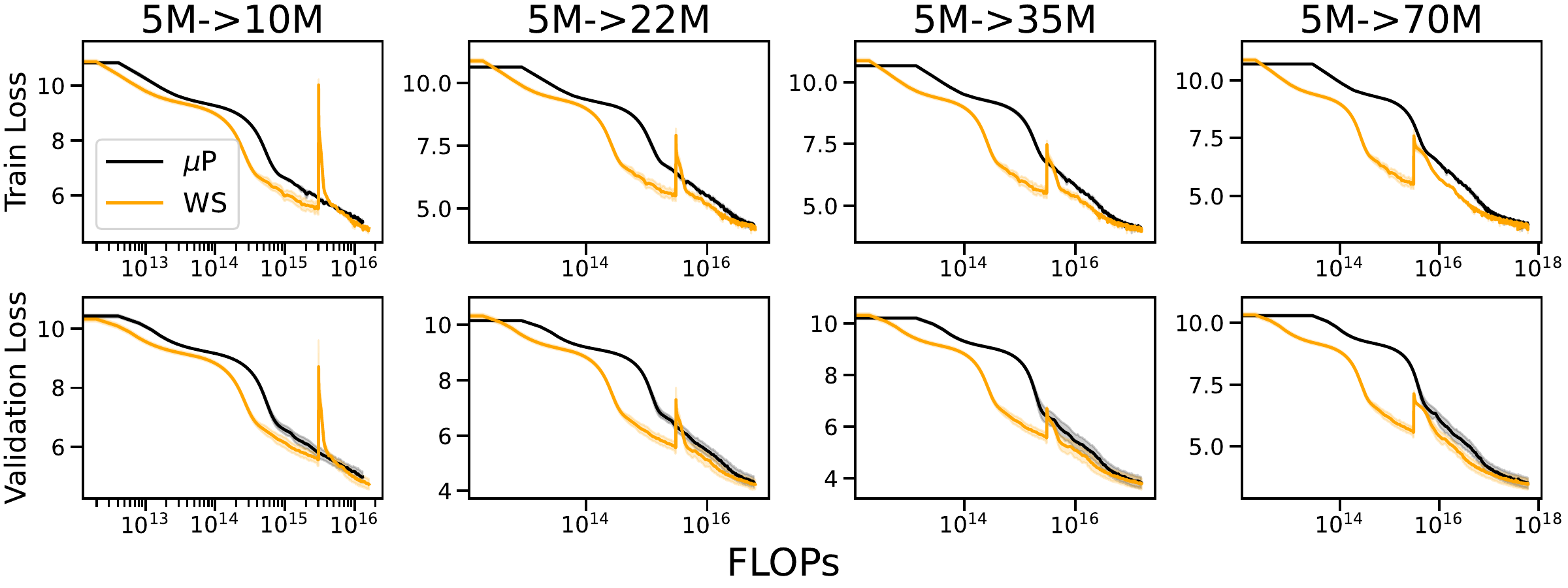} \\
        \includegraphics[width=0.75\columnwidth]{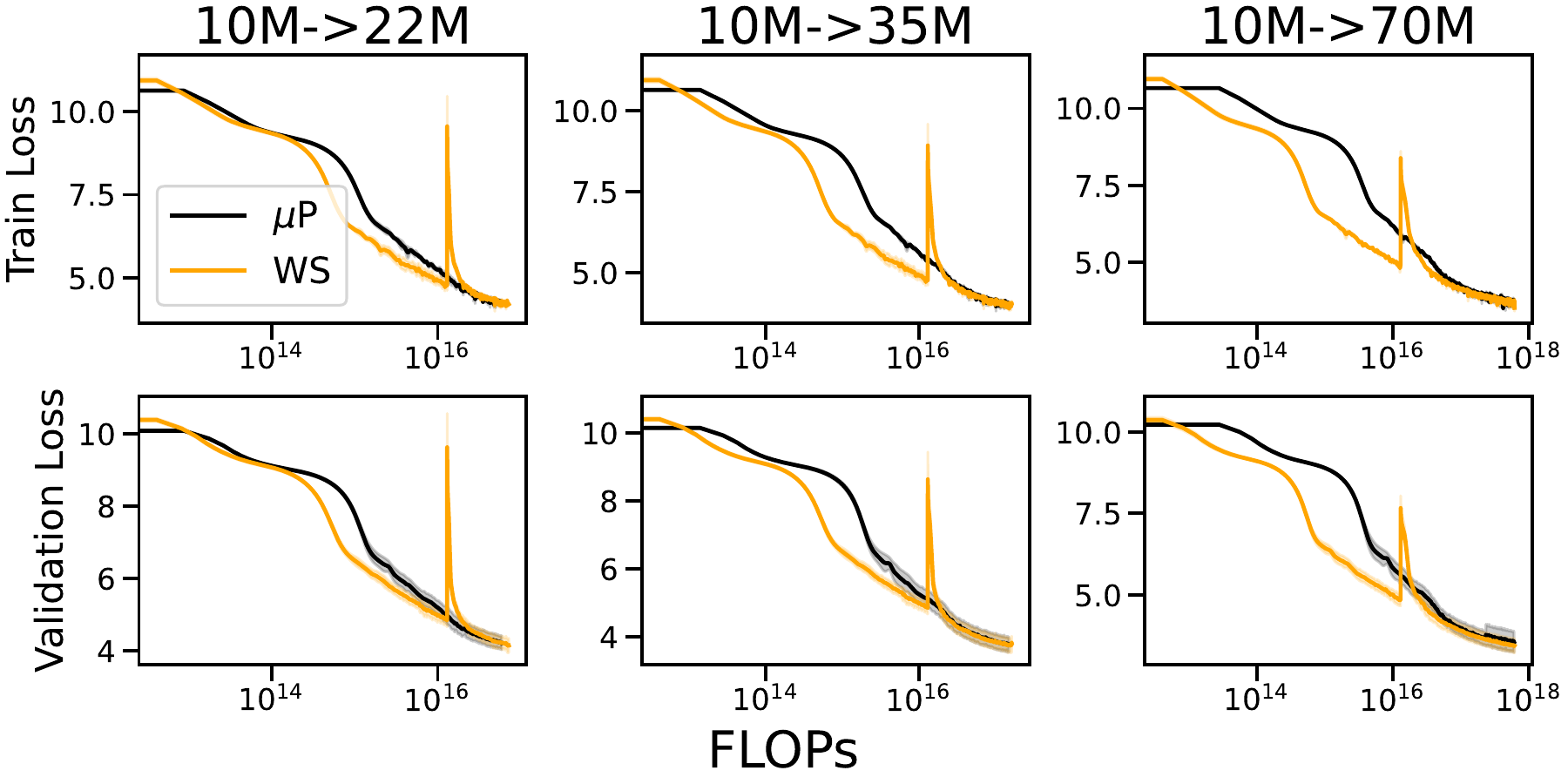} \\
        \includegraphics[width=0.5\columnwidth]{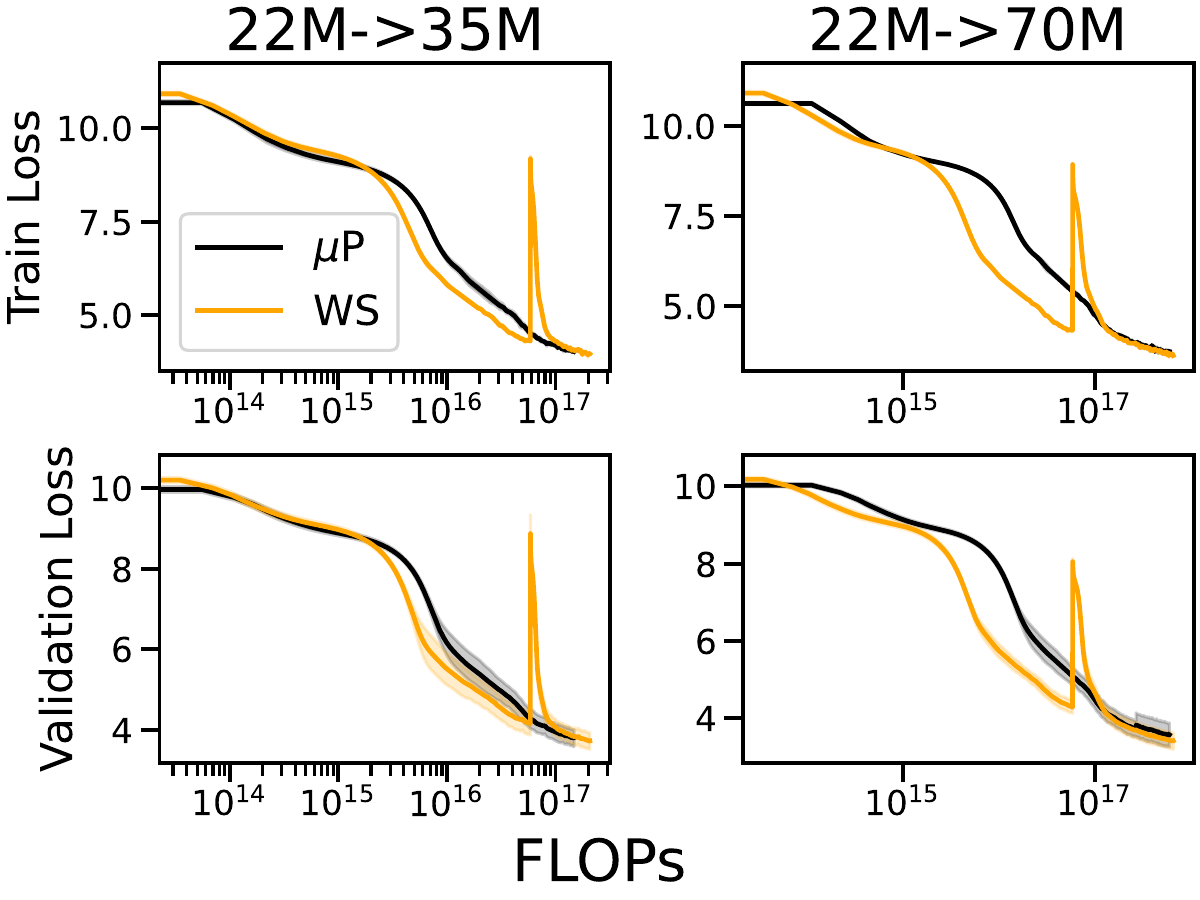} \\
    \end{tabular}
    \caption{
    Warmstarting runs with the base model learning curve included. 
    This can be seen with the position of spikes somewhat \textit{delayed} the lower the difference between the target and base model scale is.
    For correct application of $\mup{}$, in principle, the trained \textit{base} model is also available given the tuning performed to find optimal hyperparameters at the \textit{base} scale.
    }
    \label{fig:warm-mup-full-set-spike}
\end{figure}

Figure~\ref{fig:warm-longer} shows Figure~\ref{fig:warm-better-mup} when run for longer, i.e., trained with more tokens (here, 30 tokens/parameter).
We observe that the feature learning under $\mut{}$ is retained under our proposed \ws{} method.
A similar converging loss compared to vanilla-$\mup{}$ may be the limit imposed on the loss by the model scale.
Figure~\ref{fig:warm-mup-full-set} adds more \textit{base} scale transfer to our primary result, complementing Figure~\ref{fig:warm-better-mup}. 
The general trend of our \ws{} approach improving $\mup{}$ continues across all pairs of model scales we tried.
However, as expected, the extent of gains provided by \ws{} varies depending on the model scales.
For all of these results, we append the base model learning curves to the target model learning curves of our \ws{} approach in Figures~\ref{fig:warm-mup-full-set-spike}. 
Although this shifts the learning curves, the impact is minimal, particularly when there is a significant difference in size between the base and target models.

\subsection{Analyzing training dynamics}
\label{app:exp-train-dyna}

\begin{figure}
    \centering
    \includegraphics[width=\linewidth]{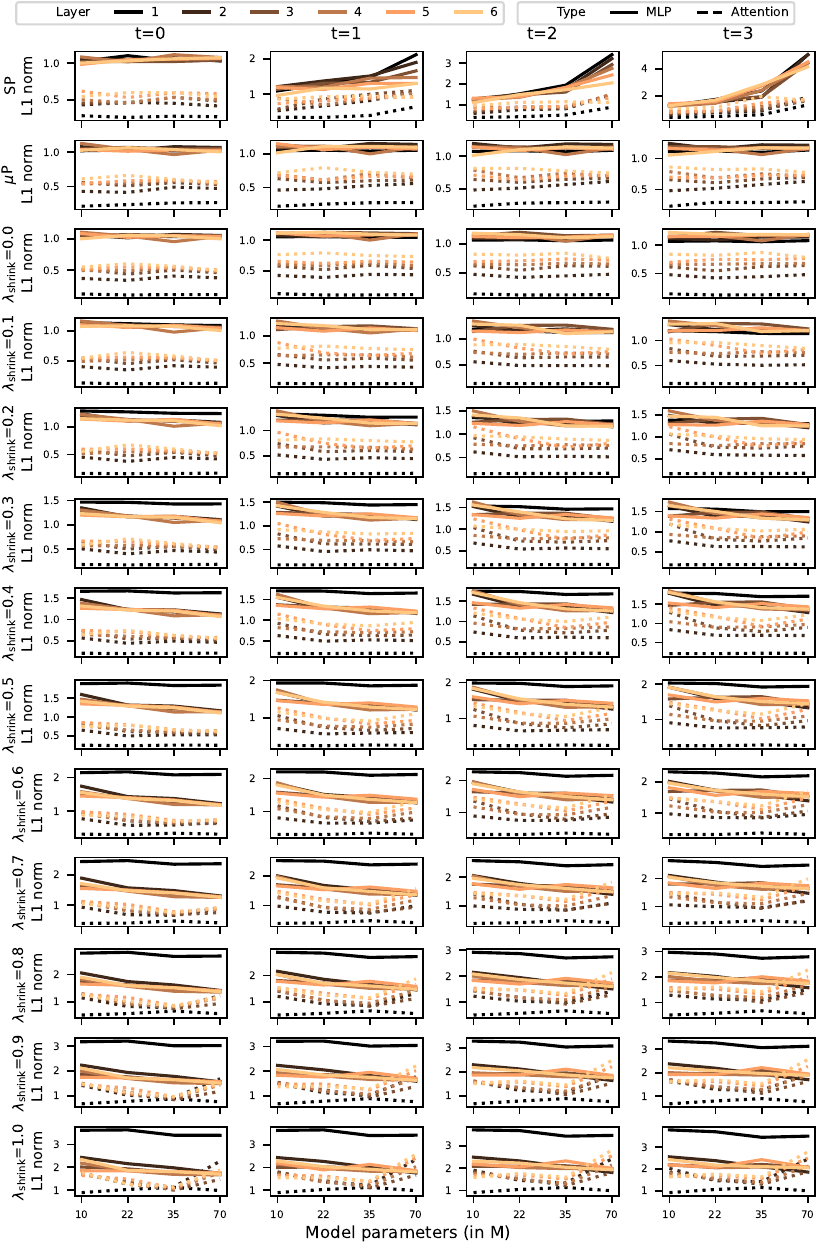}
    \caption{L1 norm of the layers activation across model scales ($x$-axis) for standard parametrization (SP), $\mup{}$ and WS with varying $\shrinkhp$ values (see, Equation~\ref{eq:warm-snp}). 
    The lesser the shrinking (higher $\shrinkhp$), the more does the instability grow with time, especially across individual layers.}
    \label{fig:coord_checks_full}
\end{figure}

\begin{figure}[htbp]
    \centering
    \begin{tabular}{c}
    \includegraphics[width=0.9\columnwidth]{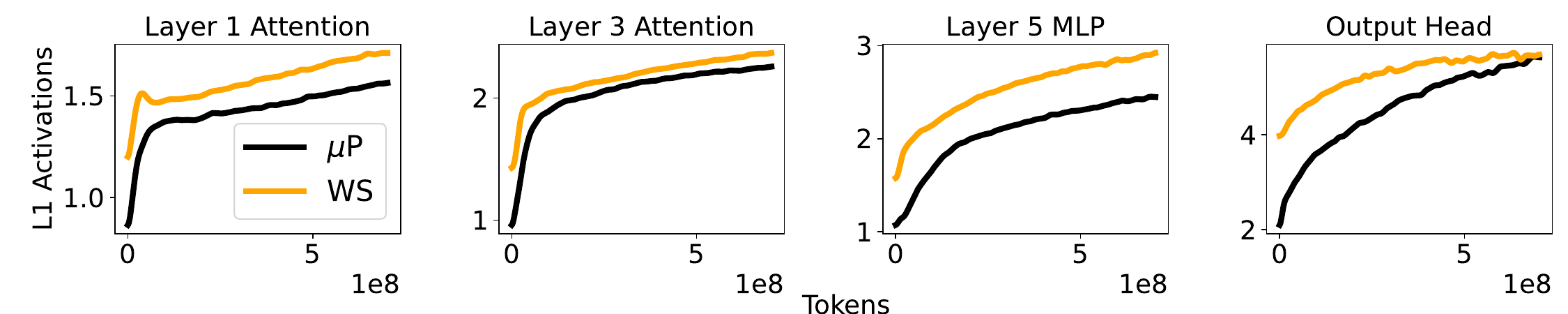}  % \includegraphics[width=\columnwidth]{neurips2024_ws/figures/warm-mup-stable.pdf}
    \end{tabular}
    \caption{
    Aggregates and extends Figure~\ref{fig:coord_checks_full}, showing L1 norm of the layers activations over the entire training run.
    Importantly warmstarting, leads to stable training, closely matching $\mup{}$ trends.}
    \label{fig:warm-mup-l1-stable}
\end{figure}

\begin{figure}[htbp]
    \centering
    \includegraphics[width=0.75\linewidth]{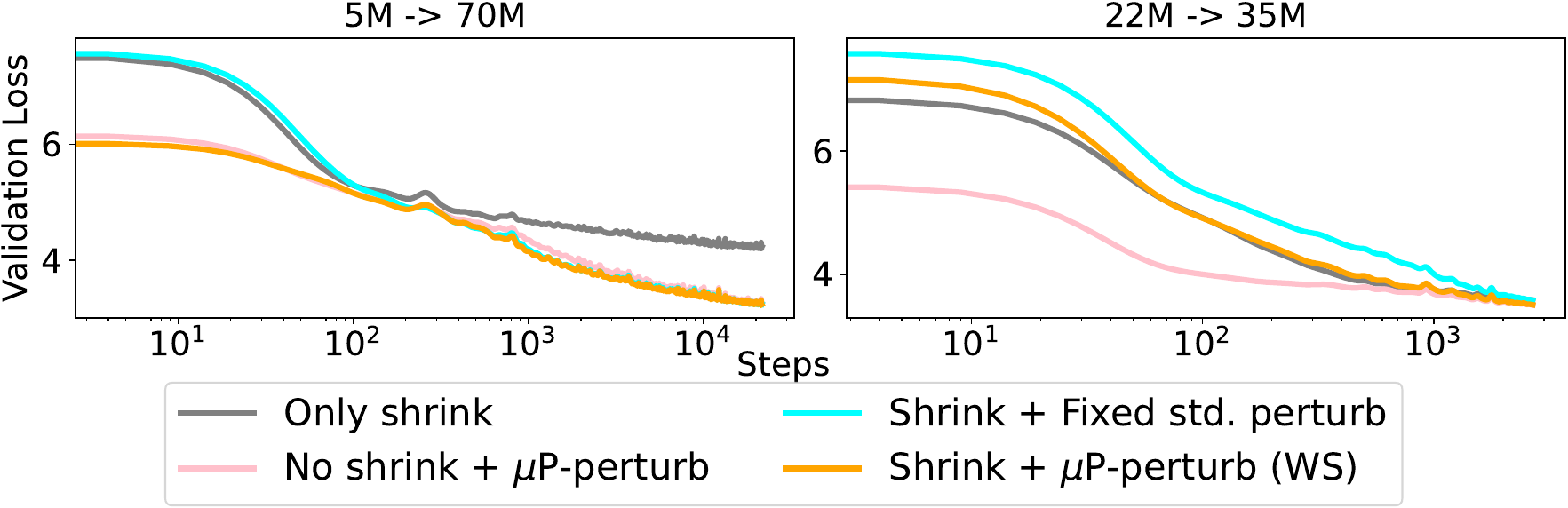}
    \caption{
    % {\color{red}Comparison of different techniques of warmstarting around the \textit{}.}
    Comparison of different techniques of warmstarting around \snp{}, to highlight the role of individual components of shrinking the base model weights and the perturbation strength.
    The relative ranking of these methods vary significantly over the relative scale differences.
    See Figure~\ref{fig:warm-shrink-check} for the corresponding plot showing the role of the magnitude of shrink.
    }
    \label{fig:snp-scale-dependent}
\end{figure}

\begin{figure}[htbp]
    \centering
    \includegraphics[width=\columnwidth]{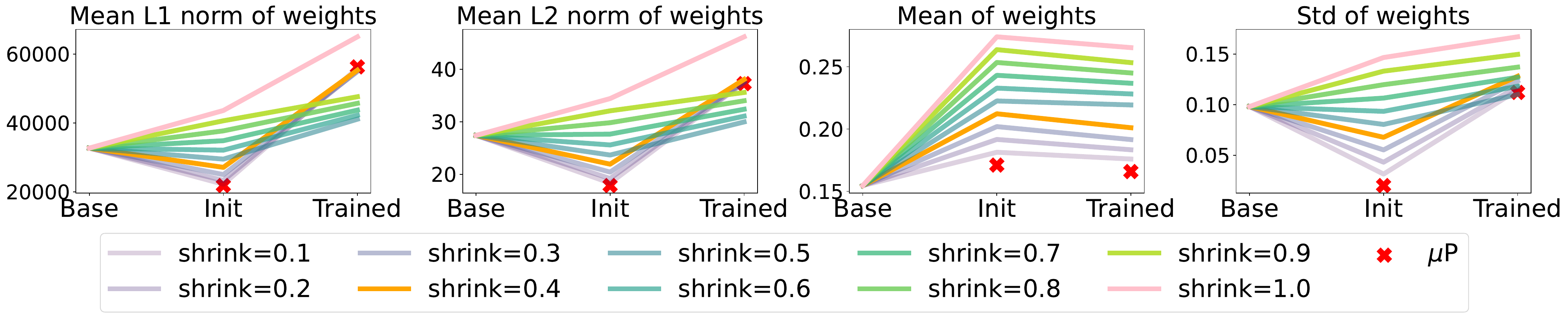}
    \caption{
    Compares the state of weights at different training times: (i) \textit{Base} indicates the weights after completion of the full training budget (here, on scale \sfour{}); (ii) \textit{Init at Warmstart} indicates the weights post-warmstarting, before beginning training (here, for scale \ssix{}); (iii) \textit{Trained} indicates the state of the warmstarted weights after expending the corresponding training budget. 
    Any higher shrinking factor ($>0.4$) interestingly leads to divergence of the final trained L1/L2 norms of the weights, especially compared to $\mup{}$.
    Note, \textit{shrink$=1.0$} corresponds to no-shrinking of weights before applying warmed-$\mup{}$.
    }
    \label{fig:warm-shrink-check}
\end{figure}

When applying \ws{} with $\mup{}$, it is important to also guarantee the optimality of the \textit{base} scale hyperparameters when scaled and transferred to the larger \textit{target} model.
This beahviour manifests in metrics that evolve over training steps, such as the L1 norm of activations, L1/L2 norms of the weights, etc.
Figures~\ref{fig:coord_checks_full},~\ref{fig:warm-mup-l1-stable},~\ref{fig:snp-scale-dependent},~\ref{fig:warm-shrink-check} together highlight that the loss perspective alone is inadequate in knowing what will work best across scales.
However, it appears that there is a \textit{sweet-spot} for $\shrinkhp$ such that it is not too low ($=0$ is equivalent to $\mup{}$) or too high ($=1$ is equivalent to not shrinking the \textit{base} model).
The value of $0.4$ suggested in the literature appears to work the best in our setup.

\subsection{Successive warmstarting}
\label{sec:successive-warms}

\begin{figure}[htbp]
\centering
\includegraphics[width=0.75\linewidth]{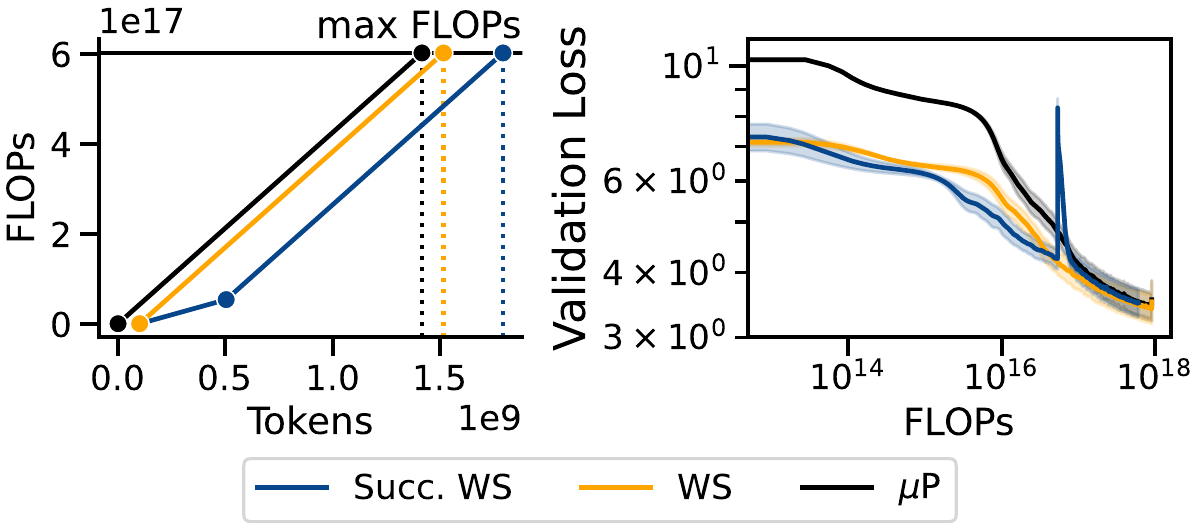}
\caption{
\textit{WS}: our approach of warmstarting-$\mup{}$ from \szero{}\textrightarrow\seight{}; 
\textit{Succ. WS}: warmstarting from \szero{}\textrightarrow\sfour{}, and then from \sfour{}\textrightarrow\seight{}, with \szero{} as the base scale for $\mut{}$ under both scale increase; 
(\textit{Left}): Under same compute allocation for training a \seight{} model, using $\mup{}$, we see that \ws{} leads to the consumption of more tokens as we do not repeat data when \ws{}, and naturally successively \ws{} further leads to more token usage;
(\textit{Right}): The learning curves for the same runs, under the same compute budget as expected for a \seight{} model. 
Despite a spike in loss (\textit{blue}) for successive warmstarting, it is able to achieve \textit{similar} loss as the vanilla-$\mup{}$ run and the warmstarted-\seight{}. 
Even though the \seight{} model in the successive warmstarting run sees much lesser compute.
}
\label{fig:succ-warm}
\end{figure}

The entire goal of warmstarting a model from a smaller model is to speed up convergence of the larger model training.
For pretraining LLMs, typically done in a sub-epoch manner by not repeating a data, we have a unique setup where if we want to train a model at a particular target scale, say $100$M, for compute-optimality~\citep{hoffmann-arxiv22a}, a warmstarted training run at $100$M will consume \textit{more tokens} than a standard run from scratch at $100$M.
This is illustrated in Figure~\ref{fig:succ-warm} (left) where we see that for the same amount of compute (in FLOPs) allocated to a vanilla-$\mup{}$ and warmstarted-$\mup{}$, the latter consumes more tokens.

Taking this further such that for the same amount of total compute spent, we start from the base model and progressively scale a model in stages.
Figure~\ref{fig:succ-warm} (left) shows in \textit{blue} how a \szero{} model was warmstarted to \sfour{} and continued training, which was then again warmstarted to \seight{}, leading to a much higher consumption in tokens.
As expected and observed (see, Figure~\ref{fig:snp-scale-dependent}) there is a spike when warmstarting and continuing a learning curve (see, Figure~\ref{fig:succ-warm} (right)).
However, the increase of model scale and our approach enables a quick drop in loss, enough for it to match $\mup{}$'s final performance.

Given that smaller models will tend to have higher loss gains per unit compute (FLOPs), this shows up in the \textit{blue} curve generally being better\footnote{Note that until the blue spike, loss of a warmstarted-\sfour{} model is better than a warmstarted-\seight{} model} than the warmstarted-\seight{} until its size is grown.
Therefore, there is potential in staging warmstarting and continually growing models for improved loss-compute or tokens-compute efficiency.
One flaw that Figure~\ref{fig:succ-warm} also highlights is that despite having \textit{spent more tokens} for the staged, successive \ws{} run, the extra tokens seen do not show up as improved loss.
Either, the compute spent in recovering from the loss spike prevents from improving the loss under total available equivalent compute.
Or, the warmstarting method is suboptimal and is under-utilizing the tokens seen overall across \textit{base} scale training and \ws{}.

We believe this is a promising direction and together with an improved \ws{} technique will lead to much more practical speed ups and offer avenues for improved hyperparameter tuning strategies. 
Moreover, studying such setups through the lens of compositional generalization~\citep{schug-arxiv24a} and \abc{}~\citep{everett-icml24a} will offer novel and practical insights and is a certain future direction to pursue.

% We can cherry pick the best base model to warmstart from and show two runs for: scale0 \textrightarrow scale4 \textrightarrow scale8.

%and scale0 \textrightarrow scale2 \textrightarrow scale4 \textrightarrow scale6 \textrightarrow scale8.

% Thus motivating \textit{successive warmstarting}.

% Highlight the compositional effect of training, data consumption, FLOPs consumption.

% \subsection{Scales as fidelity}

% {\color{red}Can we utilize our grid search from scale0, scale2, scale4 and use \textit{batch size} as a proxy for our HP correlation?}

% \note{NM: Tarek's work finds that it is likely better if batch size is increased when scaling with muP}

% Our empirical analysis investigates the relationships between GPT-2~\cite{} models across different scales. Using the GptNeoxMLP from the Lightning AI's litgpt library~\cite{}, we explore how scaling various architectural parameters impacts model performance and efficiency. Detailed specifications of each model configuration are provided in Table~\ref{tab:model_scaling}.

% \ldots
% \note{NM: Heri and I decided to drop the ablation as it was not clear why we should select 0.4 given the plot. All versions achieve the same-ish loss while \textit{Ours} is never the best in the cases shown (and worst for 22M->34M).
% It makes more sense to motivate 0.4 using the training dynamics pitch.
% }
% \begin{figure}
%     \centering
%     \includegraphics[width=0.75\linewidth]{neurips2024_ws/figures/snp_ablation.pdf}
%     \caption{Caption}
%     \label{fig:snp-ablation}
% \end{figure}

\end{document}